# Deep Bidirectional Transformers for SoC Flow Specification Mining

Md Rubel Ahmed, Hao Zheng
University of South Florida
Tampa, FL
{mdrubelahmed,haozheng}@usf.edu

## ABSTRACT

High-quality system-level message flow specifications can lead to comprehensive validation of system-on-chip (SoC) designs. We propose a disruptive method that utilizes attention mechanism to produce accurate flow specifications from SoC IP communication traces. The proposed method can overcome the inherent complexity of SoC traces induced by the concurrency and parallelism of multicore designs that existing flow specification mining tools often find extremely challenging. Experiments on highly interleaved traces show promising flow reconstruction compared to several tools dedicated to the flow specification mining problem.

## 1 INTRODUCTION

Existing pattern or flow mining tools performs poorly on system level flow specification mining for SoC traces due to the highly concurrent nature of SoC executions. Development in SoC trace mining for producing flow specifications faces an extremely difficult challenges when the SoC design complexity increases with the addition of multiple cores, multiple level of shared and private caches and high-performance interconnects. During execution of an SoC design, instances of flows it implements are executed concurrently. Typically, multiple instances of different flows that are executed concurrently are captured in the traces. We use the term message to denote the unit of exchange between two IP in the SoC. Though complex in nature, transaction-level traces are result of flow executions which could be represented as directed accyclic graph (DAG) in the SoC architectural document. However, as the SoC design life-cycle advances, such flow specifications are not generally updated or becomes impossible keep updating manually. Lack of flow specification hinders effective validation and analysis of SoC design.

Several works have addressed this so called specification mining problem both from hardware and software domains [1, 2, 6]. As the traces are orderly arrangements of events or messages that follows some ground truths we want to know, attention mechanism in sequence modeling comes with a big promise in this research [4]. In this paper, we present a specification mining framework, to address the specification mining from highly concurrent traces. The overview of our flow mining method is shown in Figure 1. It takes as input a set of execution traces over messages observed in various communication links in an SoC design, and produces a set of flows that can describe the traces. A lightweight masked language model (LM) encoder is trained on the traces that captures the context of the input traces in the form of latent space. This model acts as an next word predictor during a causality graph guided search to find each flow from the causality graph whenever a branching decision to be made for the respective flow.

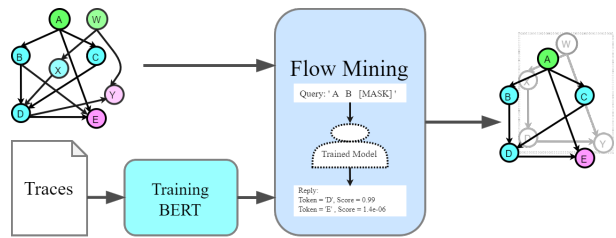

Figure 1: The overview of Flow Mining from SoC execution traces. Messages from traces are used to build the causality graph which is also a DAG. DistilBERT [5] is trained on the traces. Causality graph is then refined to the form of message flow specification for each $(Start, End)$ message pairs with the help a trained model.

## 2 MINING FRAMEWORK

Message flow specifications is a protocol of how the system-level functions are implemented on the SoC. Each flow specification is a collection of messages as nodes and edges as their temporal or causal relation expressed as a DAG. The flow specification for an SoC is a set of flows, denoted as $\vec{F} = \{F_i\}$. Each message is a triple (src : dest : cmd) where the src denotes the originating component of the message while dest denotes the receiving component of the message. Field cmd denotes the operation to be performed at dest. Each message flow is associated with an unique start message, and may terminate with one or multiple different messages. During execution of an SoC design, instances of flows it implements are executed concurrently. A SoC execution trace is a sequence of sets of message instances. An example trace resulting from executing each path in the flows in Figure 1 (top left) once where both flows are instantiated (A, B, D, W, X, D, Y, E). Messages (A, E) and (W, Y) are $Start$ and $End$ message pairs. The flow specifications in Figure 1 also define causality relations among messages which is same as in *FlowMiner* [1]. Messages $m_i$ and $m_j$ satisfy the structural **causality** property, denoted as $causal(m_i, m_j)$, if $m_i.\text{src} = m_j.\text{dest}$

Algorithm 1 implements the mining framework shown in Figure 1. It consists of three major functions Trace Processing, Encoder Training, Flow Mining, which are described below.

**Trace Processing** Proposed framework accepts transaction-level traces as sequences of messages as input. It assumes that source

**Algorithm 1: Flow Mining**

1: **Input:** a set of traces $\rho$, probability threshold $\theta$
2: **Output:** a set of message flows $\mathcal{M}$
3: Extract unique messages in $\rho$ into $M$;
4: Build the causality graph $G$ from $M$;
5: Train BERT $B$ on $\rho$;
6: **foreach** $(Start, End) \in G$ **do**
7:   Query $B$ for next message $m.score() \geq \theta$ towards $End$;
8:   Return causality subgraph as a $Flow_{(Start, End)}$;

and destination information of each message are known and messages are observed in between IP communication fabrics. A causality graph $G$ is constructed from the messages that contains all the flows exercised in the execution. For every traversal path $s = (m_0, m_1, \ldots, m_n)$ in $G$, the following condition must hold.

$$\forall 0 \leq i \leq n-1, causal(m_i, m_{i+1}) \text{ is true.}$$

Trace preparation for training also preserves the causality. The proposed model reads one line at a time, therefore traces are divided into smaller parts or traces $\rho_t \subseteq \rho$.

**Encoder Training** We utilize Masked LM *DistilBertForMaskedLM* model keeping the original architecture intact and retrain it on the traces to predict masked token in a sequence. We use the light version of BERT [3] namely *DistilBERT* [5] which is only 6 layers. This bidirectional model is exceptionally powerful than the shallow left-to-right or otherwise models. The objective is to enable the model to predict next word in a multi-layered context. To train the deep bidirectional encoder representation, we mask a certain percentage of the input traces in token form, randomly. We apply several fine-tuning such as: making the model read traces line-by-line, changing mask probability (default is 15%) and learning rates to fit our traces better.

**Flow Mining** Once the model is trained, the flow mining is straight forward. Mining happens in the causality graph guided fashion. The initial causality graph for each flow might have some illegal messages or dependency added just because a message is causal to another. A guided search is starts from the start message and reaches to the end message forming a causality sub-graph $G_f \subseteq G$. The trained BERT is able to predict if two messages $m_1$ and $m_2$ though satisfy the condition $causal(m_1, m_2)$ should be considered in the flow graph. Therefore, if the model returns $m_2$ for $m_1$ with a greater than the threshold probability, we keep them in the graph, otherwise, they are ignored. This process is repeated until all flows in $\vec{F}$ is discovered for each start, end message pairs.

## 3 EXPERIMENTAL RESULTS

The strength of this approach is that flows can be produced from highly concurrent interleaved traces. A good method for evaluating produced flows could be having some ground truth flows for similarity comparison. Therefore, to demonstrate the capability, we select two flows from a standard SoC design and synthetically generate execution traces. Two flows we chose represent `CPU0_Read` and `UART_Upstream_Read` each of which has 3 branches, 14 unique messages, and 4 shared messages as in Figure 2 (left). A training set is prepared containing 600 concurrent and interleaved runs of these flows. We train the model for 10 epochs which took around 6.00 hrs. We set the masking a token probability is set to 30%.

**Table 1: Mining results for the three methods on evaluation metrics described in *FlowMiner* [1]. $\theta$ = 0.75 for flow mining.**

|  | #Patterns Mined | Precision | Recall |
| --- | --- | --- | --- |
| *FlowMiner* [1] | 55 | 100% | 25% |
| *LSTM* [2] | 24 | 100% | 3.12% |
| *Ours* | All | 100% | 100% |

Figure 2 shows the mined flow for `CPU0_Read`. A unrefined version of this flow will have an additional branch travelling `Message_27` or for simplicity message 27 in the following discussion. The general understanding is that all the branches will originate from the *start* message and terminate in *end* message. Therefore, branches that does not lead to 26 is discarded in the first place. After that, a query to the BERT for the next message after 19 does not predict 27 for the given answer probability threshold of 75%. Therefore, we discard this message. This process continues till we examine every branches leading to the end message 26.

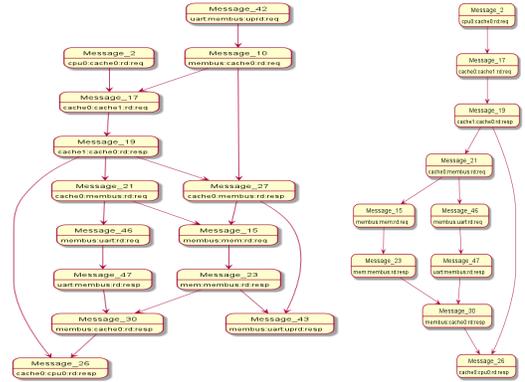

**Figure 2: Causality graph for `CPU0_Read` and `UART_Upstream_Read` flows (left). Former flow starts with `Message_2` and ends with `Message_26` that BERT refines (right) by removing illegal `Message_27` which resembles the actual `CPU0_Read` exercised in the trace.**

We also compare with an LSTM based pattern mining approach [2] and a data mining approach [1]. Work [2] mines flow specification in the form of sequential patterns as flows can be viewed as a collection of patterns representing all paths in a particular flow. We extract patterns for different values of pattern probability, but could not produce complete specification for any of the test flows in execution. Table 1 summaries results for this tool on the training traces. Pattern mining approach [1] utilizes association rule mining. This method finds invariants from the traces. The mined pattern can partially construct `CPU0_Read` where message 15 and 23 are missing. In terms of evaluation metrices: recall and precision described in *FlowMiner*, proposed method beats both of them.

We find causality graph driven encoder based next word prediction model produces best quality flows specifications. The training time could be a limiting factor for this approach than existing approaches. Attribute based slicing techniques could help reduce the training time in this regard.